\documentclass[english,a4paper,conference]{IEEEtran}
\usepackage[T1]{fontenc}
\usepackage[latin9]{inputenc}
\usepackage{color}
\usepackage{array}
\usepackage{multirow}
\usepackage{amsbsy}
\usepackage{amstext}
\usepackage{amssymb}
\usepackage{graphicx}

\makeatletter

\providecommand{\tabularnewline}{\\}

\usepackage{algorithmicx}
\usepackage{microtype}

\date{}

\makeatother

\usepackage{babel}
\begin{document}

\title{Knowledge Graph Embedding with Multiple Relation Projections}

\author{Kien Do, Truyen Tran, Svetha Venkatesh\\
Applied AI Institute, Deakin University\\
\emph{\{dkdo,truyen.tran,svetha.venkatesh\}@deakin.edu.au}}
\maketitle
\begin{abstract}
Knowledge graphs contain rich relational structures of the world,
and thus complement data\textendash driven machine learning in heterogeneous
data. One of the most effective methods in representing knowledge
graphs is to embed symbolic relations and entities into continuous
spaces, where relations are approximately linear translation between
projected images of entities in the relation space. However, state-of-art
relation projection methods such as TransR, TransD or TransSparse
do not model the correlation between relations, and thus are not scalable
to complex knowledge graphs with thousands of relations, both in computational
demand and in statistical robustness. To this end we introduce TransF,
a novel translation\textendash based method which mitigates the burden
of relation projection by explicitly modeling the basis subspaces
of projection matrices. As a result, TransF is far more light weight
than the existing projection methods, and is robust when facing a
high number of relations. Experimental results on the canonical link
prediction task show that our proposed model outperforms competing
rivals by a large margin and achieves state-of-the-art performance.
Especially, TransF improves by 9\%/5\% in the head/tail entity prediction
task for N-to-1/1-to-N relations over the best performing translation-based
method.

\end{abstract}
\global\long\def\Neigh{\mathcal{N}}
\global\long\def\Real{\mathbb{R}}
\global\long\def\Loss{\mathcal{L}}
\global\long\def\Relations{\mathcal{R}}
\global\long\def\Triplets{\mathcal{T}}
\global\long\def\Entities{\mathcal{E}}
\global\long\def\Xcal{\mathcal{X}}
\global\long\def\Ncal{\mathcal{N}}
\global\long\def\Arm{\mathrm{A}}
\global\long\def\Nrm{\mathrm{N}}
\global\long\def\Erm{\mathrm{E}}
\global\long\def\Ebm{\boldsymbol{\mathrm{E}}}
\global\long\def\Rbm{\boldsymbol{\mathrm{R}}}
\global\long\def\Rrm{\mathrm{R}}
\global\long\def\Trm{\mathrm{T}}
\global\long\def\arm{\mathrm{a}}
\global\long\def\drm{\mathrm{d}}
\global\long\def\srm{\mathrm{s}}
\global\long\def\frm{\mathrm{f}}
\global\long\def\hbm{\boldsymbol{\mathrm{h}}}
\global\long\def\tbm{\boldsymbol{\mathrm{t}}}
\global\long\def\rbm{\boldsymbol{\mathrm{r}}}
\global\long\def\ebm{\boldsymbol{\mathrm{e}}}
\global\long\def\pbm{\boldsymbol{\mathrm{p}}}
\global\long\def\ubm{\boldsymbol{\mathrm{u}}}
\global\long\def\wbm{\boldsymbol{\mathrm{w}}}
\global\long\def\alphab{\boldsymbol{\alpha}}
\global\long\def\betab{\boldsymbol{\beta}}
\global\long\def\bbm{\boldsymbol{\mathrm{b}}}
\global\long\def\Ibm{\boldsymbol{\mathrm{I}}}
\global\long\def\Wbm{\boldsymbol{\mathrm{W}}}
\global\long\def\Wbcal{\boldsymbol{\mathcal{W}}}
\global\long\def\Mbm{\boldsymbol{\mathrm{M}}}
\global\long\def\Ubm{\boldsymbol{\mathrm{U}}}
\global\long\def\Vbm{\boldsymbol{\mathrm{V}}}

\section{Introduction}

Current data\textendash driven machine learning works well in an homogeneous
domain, but may not scale to domains that demand heterogeneous knowledge
about entities and relations. Knowledge graphs (KGs), which encompass
\textcolor{black}{rich information about structures of the world},
offer a complementary approach. Knowledge\textendash augmented machine
learning thus holds a promise to improve performance through knowledge
reuse and to enable explanability \cite{wilcke2017knowledge}. These
benefits have been found in many applications, ranging from vision
\cite{marino2016more}, recommendation \cite{chaudhari2017entity},
question answering \cite{bordes2015large,miller2016key,das2017question}
to language modeling \cite{ahn2016neural}. Despite huge efforts spent
to build large-scale KGs such as Freebase \cite{bollacker2008freebase},
YAGO \cite{suchanek2007yago} or DBpedia \cite{auer2007dbpedia},
a major problem consistently remains: they are far from complete.
Thus, it poses a canonical task of automatic completion from the existing
knowledge base, which amounts to reasoning about unknown relations
between entities. 

A typical KG is represented as a graph whose nodes are entities and
edges are relations between heads and tails. While this raw representation
is adequate to store known knowledge, relating distant entities requires
expensive graph traversal, possibly through multiple paths. Thus knowledge
graph completion calls for learning of a new representation that supports
scalable reasoning. The most successful approach thus far is through
embedding entities and relations into a continuous vector space, which
naturally lends itself to simple algebraic manipulations \cite{bottou2014machine}.
A well known method is TransE \cite{bordes2013translating}, which
embeds entities and relations into the same space where the difference
between head and tail is approximately the relation. While this embedding
permits very simple translation-based relational inference, it is
too restrictive in dealing with 1-to-N, N-to-1 and N-to-N relations.

An effective solution is to consider two separate embedding spaces
for entities and relations. Entities are then mapped into the relation
space using \emph{relation\textendash specific projections}, such
as those in TransR \cite{lin2015learning}. This mapping strategy,
however, causes critical drawbacks. First, when the number of relations
is large, the whole projection matrices are expensive to model. Second,
treating each relation separately does not account for the latent
structure in the relation space, leading to waste of resources. An
example of such a latent structure is the correlation between relations
``nationality'' and ``place-of-birth'', as the latter may infer
about the former. 

\begin{figure}
\begin{centering}
\includegraphics[width=0.9\columnwidth]{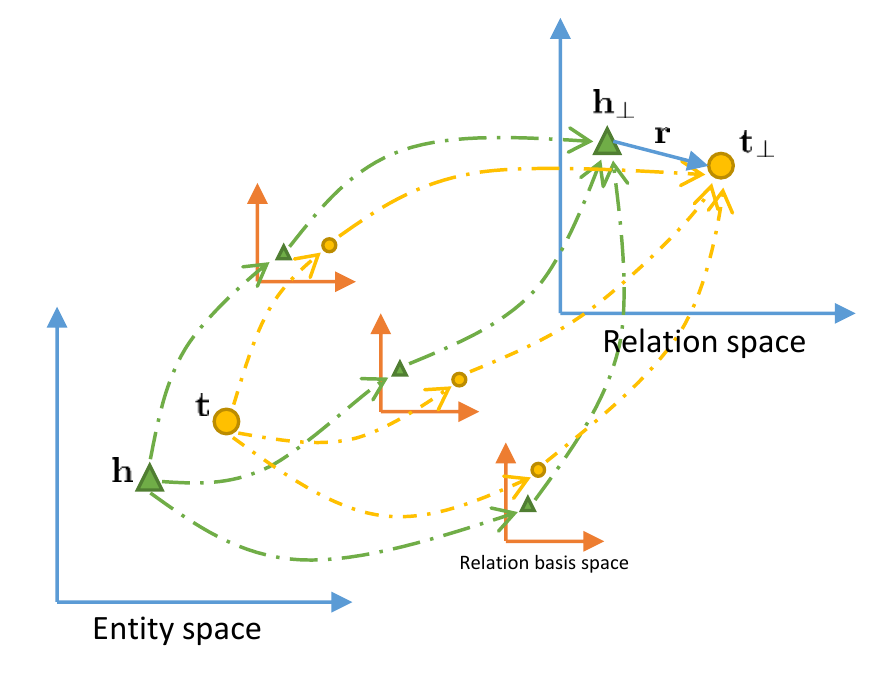}
\par\end{centering}
\caption{Illustration of TransF where head embedding $\protect\hbm$ and tail
embedding $\protect\tbm$ are projected into multiple subspaces before
combined by relation\textendash specific coefficients into the final
vectors $\protect\hbm_{\perp}$ and $\protect\tbm_{\perp}$ in the
relation space. In this space, translation\textendash based reasoning
is assumed as in TransE \cite{bordes2013translating}, i.e., $\protect\hbm_{\perp}+\protect\rbm\approx\protect\tbm_{\perp}$.
\label{fig:TransRF}}
\end{figure}

To this end we propose a new translation-based method called TransF,
which is inspired by TransR, but does not suffer from these problems.
Under TransF, projection matrices are members of a matrix space spanned
by a fixed number of matrix bases. A relation\textendash specific
projection matrix is characterized by a relation\textendash specific
coordinate in the space. Put in other way, the relation projection
tensor is factorized into product of a relation coordinate matrix
and a basis tensor. Hence, TransF is much more efficient and robust
than TransR. Fig.~\ref{fig:TransRF} illustrates the idea behind
TransF. 

We evaluate our TransF on the common link prediction task using two
popular KGs: Freebase and WordNet. The experimental results show that
TransF delivers significant improvements over state-of-the-art translation\textendash based
methods.

\section{Preliminaries}

Let us first define common notations. A knowledge graph (KG) is constructed
from a set of entities $\Entities$ and a set of relations $\Relations$.
The basic unit representing a fact in KG is a triple, denoted as $(h,r,t)$
where $h$ is head entity, $r$ is relation and $t$ is tail entity.
We use bold letters with normal font to indicate vectors and bold
letters with capital font to indicate matrices. In this fashion, $\hbm,\rbm,\tbm$
are the embedding vectors of $h,r,t$, respectively.

\paragraph{TransE:}

TransE \cite{bordes2013translating} is a simple yet scalable and
effective method for knowledge graph representation. It treats entities
as points and \emph{relations as translation vectors} in the same
embedding space $\Real^{\drm}$. Thus relational reasoning between
any entities is straightforward. The energy function of a particular
triple $(h,r,t)$ is defined as:

\[
E(h,r,t)=\|\hbm+\rbm-\tbm\|_{\ell_{1/2}}
\]
The energy should be low for true triples and high for incorrect triples.
TransE performs well when relations are 1-to-1 but fails to handle
relations of type 1-to-N, N-to-1 or N-to-N \cite{bordes2013translating}.
For example, if $r$ is an N-to-1 relation in which $(h_{1},r,t)$,
$(h_{2},r,t)$,..., $(h_{n},r,t)$ hold, minimizing the energy function
will force $\hbm_{1}\approx\hbm_{2}\approx...\approx\hbm_{n}$. It
is undesirable since $h_{1},h_{2},...,h_{n}$ are different entities.

\paragraph{TransH:}

The main drawback of TransE is that an entity exhibits identical characteristic
despite involving in different relations. To solve this problem, TransH
\cite{wang2014knowledge} projects head and tail entities onto a relation\textendash specific
hyperplane before doing translation as follows:

\[
\hbm_{\perp}=\hbm-\wbm_{r}^{\intercal}\hbm\wbm_{r},\ \ \ \tbm_{\perp}=\tbm-\wbm_{r}^{\intercal}\tbm\wbm_{r}
\]
The energy function now becomes:

\begin{equation}
E(h,r,t)=\|\hbm_{\perp}+\rbm-\tbm_{\perp}\|_{\ell_{1/2}}\label{eq:TransH_energy}
\end{equation}
where $\hbm_{\perp},\tbm_{\perp}\in\Real^{\drm}$ are head and tail
projected vectors, respectively; and $\wbm_{r}\in\Real^{\drm}$ is
the unit-length normal vector of the hyperplane $\mathrm{H}_{r}$
with respect to $r$. 

\paragraph{TransR:}

Since entities and relations are separate objects, it is intuitive
to represent them in distinct spaces. TransR \cite{lin2015learning}
implements this idea by using a relation projection matrix $\Mbm_{r}\in\Real^{\drm_{r}\times\drm_{e}}$
to map entity embeddings into the relation space:

\[
\hbm_{\perp}=\Mbm_{r}\hbm,\ \ \ \tbm_{\perp}=\Mbm_{r}\tbm
\]
TransR uses the same energy function as TransH but the translation
vector $\rbm$ is defined in the relation space instead of the relation
hyperplane. However, TransR introduces a huge number of additional
parameters, causing poor robustness in learning and scalability issues
for large KGs.

\section{Our Method: TransF}

\subsection{Correlation Among Relations.}

In several knowledge bases such as Freebase \cite{bollacker2008freebase},
relations are organized in hierarchies. Relations belonging to the
same subtree (e.g. containing similar prefixes) are correlated. Most
relationships come from the semantic meaning of relations and some
are partially independent of the entities. For example, in the FB15k
dataset, two relations \textit{``/people/person/.../major\_field\_of\_study''}
and \textit{``/people/person/.../degree''} are highly dependent
as the field of study is usually the main factor that affects the
degree one can receive (e.g. Computer Science $\rightarrow$ Bachelor
of Computer Science). Note that, in this case, we do not need any
information about the head entity (a particular person) to decide
this correlation. Another example is the pair of relations ``\textit{/people/person/place\_of\_birth}''
and ``\textit{/people/person/nationality}'' where a person born
in New York is likely to have the United States nationality. Therefore,
we argue that better modeling of correlations among relations will
lead to more accurate representation of knowledge graphs.

\subsection{TransF.\label{subsec:TransRF}}

Motivated from the above observation, we now present TransF, a new
translation\textendash based embedding method of knowledge graphs.
Our model defines two factorized projection matrices as follows:

\begin{eqnarray}
\Mbm_{r,h} & = & \sum_{i=1}^{\srm}\alpha_{r}^{(i)}\Ubm^{(i)}+\Ibm\label{eq:Mrh}\\
\Mbm_{r,t} & = & \sum_{i=1}^{\srm}\beta_{r}^{(i)}\Vbm^{(i)}+\Ibm\label{eq:Mrt}
\end{eqnarray}
where $\srm\in\Real$ is the number of factors; $\Ubm^{(i)},\Vbm^{(i)}\in\Real^{\drm_{e}\times\drm_{r}}$
$\forall i=\overline{1,\srm}$ are relation space bases for projecting
head and tail entities, respectively; $\alpha_{r}^{(i)}$ and $\beta_{r}^{(i)}$
are the corresponding coefficients of $\Ubm^{(i)}$ and $\Vbm^{(i)}$
characterized by the relation $r$. $\Ibm$ is the identity matrix,
which serves as the base case when projections are not needed. It
also provides a way to initialize TransF from TransE.

With this formulation, the underlying relationships among relations
are explicitly encoded in the relation space bases $\{\Ubm^{(i)}|i=\overline{1,\srm}\}$
and $\{\Vbm^{(i)}|i=\overline{1,\srm}\}$. $\alpha_{r}^{(i)}$ and
$\beta_{r}^{(i)}$, on the other hand, adapt this commonness to specific
relations. To deal with 1-to-N and N-to-1 relations, we project head
and tail entities into spaces described by $\Mbm_{r,h}$ and $\Mbm_{r,t}$,
respectively: 

\[
\hbm_{\perp}=\Mbm_{r,h}\hbm,\ \ \ \tbm_{\perp}=\Mbm_{r,t}\tbm
\]

Since $\Mbm_{r,h}$ and $\Mbm_{r,t}$ are factorized, each entity
is now associated with multiple yet specific views rather than a single
general view as in TransR. The energy function is similar to Eq.~\ref{eq:TransH_energy}.
To avoid trivial solution (e.g. all embedded vectors are zeros) when
minimizing the energy, we impose the following constraints in our
model: $\|\hbm_{\perp}\|_{2}=\|\tbm_{\perp}\|_{2}=1$, $\|\rbm\|_{2}\leq1$.

\paragraph{Training loss:}

We define the following margin-based loss function:

\[
\Loss=\sum_{(h,r,t)\in\Triplets}\sum_{(h',r',t')\in\Triplets'}\left[E(h,r,t)+\gamma-E(h',r',t')\right]_{+}
\]
where $[x]_{+}$ denotes $\text{max}(0,x)$, $\gamma$ is the margin,
$\Triplets$ is the set of positive triples and $\Triplets'$ is the
set of negative triples. Since the collection of all negative triples
is huge, we need to focus on negative triples which are close to the
correct ones. Similar to \cite{bordes2013translating}, given each
correct triple $(h,r,t)$ sampled during training, we generate a negative
example by replacing either head or tail entities but not both. Thus,
the form of $\Triplets'$ is: 

\[
\Triplets'=\{(h',r,t)|h'\in\Entities\}\cup\{(h,r,t')|t'\in\Entities\}\ \ \ \forall(h,r,t)\in\Triplets
\]

We apply ``bern'' sampling trick suggested in \cite{wang2014knowledge}
when corrupting triples to reduce the false-negative rate. In this
setting, the probability of sampling head and tail entities are not
equal but depends on relation types. For each relation $r$, denote
$\text{hpt}$ as the average number of head entities per tail entity
and $\text{tph}$ as the number of tail entities per head entity.
Then, given a triple $(h,r,t)$, the probability of corrupting $h$
and $t$ is $\frac{\text{tph}}{\text{tph}+\text{hpt}}$ and $\frac{\text{hpt}}{\text{tph}+\text{hpt}}$,
respectively.

\section{Experiments and Results}

\subsection{Datasets.}

For our experiments, we use two common datasets FB15k and WN18 \cite{bordes2013translating}
and their corresponding updated version FB15k-237 \cite{toutanova2015observed}
and WN18RR \cite{dettmers2017convolutional}. According to \cite{dettmers2017convolutional},
FB15k-237 and WN18RR do not contain reversible relations like FB15k
and WN18, thus, are more difficult for link prediction task. Statistics
of the datasets are provided in Table.~\ref{tab:dataset}.

\begin{table}
\caption{Statistics of datasets used in the experiments.\label{tab:dataset}}

\centering{}%
\begin{tabular}{|c|c|c|c|c|c|}
\hline 
\multirow{2}{*}{{\scriptsize{}Dataset}} & \multirow{2}{*}{{\scriptsize{}\#Train}} & \multirow{2}{*}{{\scriptsize{}\#Valid}} & \multirow{2}{*}{{\scriptsize{}\#Test}} & \multirow{2}{*}{{\scriptsize{}\#Ent}} & \multirow{2}{*}{{\scriptsize{}\#Rel}}\tabularnewline
 &  &  &  &  & \tabularnewline
\hline 
\hline 
{\scriptsize{}FB15k} & {\scriptsize{}483,142} & {\scriptsize{}50,000} & {\scriptsize{}59,071} & {\scriptsize{}14,951} & {\scriptsize{}1,345}\tabularnewline
\hline 
{\scriptsize{}FB15k-237} & {\scriptsize{}272,115} & {\scriptsize{}17,535} & {\scriptsize{}20,466} & {\scriptsize{}14,541} & {\scriptsize{}237}\tabularnewline
\hline 
{\scriptsize{}WN18} & {\scriptsize{}141,442} & {\scriptsize{}5,000} & {\scriptsize{}5,000} & {\scriptsize{}40,943} & {\scriptsize{}18}\tabularnewline
\hline 
{\scriptsize{}WN18RR} & {\scriptsize{}86,835} & {\scriptsize{}3,3034} & {\scriptsize{}3,134} & {\scriptsize{}40,943} & {\scriptsize{}11}\tabularnewline
\hline 
\end{tabular}
\end{table}

\subsection{Link Prediction.\label{subsec:Link-Prediction}}

\begin{table*}
\caption{Link prediction results on WN18, FB15k, WN18RR and FB15k-237. \protect \\
{*}: The result of DistMult and ComplEx on WN18RR and FB15k-237 are
taken from \cite{dettmers2017convolutional}.\label{tab:link_pred}}

\centering{}%
\begin{tabular}{|c|c|c|c||c|c|c||c|c|c||c|c|c|}
\hline 
\multirow{2}{*}{Method} & \multicolumn{3}{c||}{WN18} & \multicolumn{3}{c||}{FB15k} & \multicolumn{3}{c||}{WN18RR} & \multicolumn{3}{c|}{FB15k-237}\tabularnewline
\cline{2-13} 
 & MR & MRR & \multicolumn{1}{c||}{Hits@10} & MR & MRR & \multicolumn{1}{c||}{Hits@10} & MR & MRR & Hits@10 & MR & MRR & Hits@10\tabularnewline
\hline 
TransE & 251 & - & 89.2 & 125 & - & 47.1 & - & - & - & - & - & -\tabularnewline
TransH & 388 & - & 82.3 & 87 & - & 64.4 & - & - & - & - & - & -\tabularnewline
TransR & 225 & - & 92.0 & 77 & - & 68.7 & - & - & - & - & - & -\tabularnewline
CTransR & 218 & - & 92.3 & 75 & - & 70.2 & - & - & - & - & - & -\tabularnewline
TransD & 212 & - & 92.2 & 91 & - & 77.3 & - & - & - & - & - & -\tabularnewline
TransSparse (s) & 221 & - & 92.8 & 82 & - & 79.5 & - & - & - & - & - & -\tabularnewline
TransSparse (us) & 211 & - & 93.2 & 82 & - & 79.9 & - & - & - & - & - & -\tabularnewline
\hline 
TransF & \textbf{198} & 0.856 & \textit{95.3} & 62 & 0.564 & \textit{82.3} & \textbf{3246} & \textbf{0.505} & 49.8 & \textbf{210} & 0.286 & \textbf{47.2}\tabularnewline
\hline 
PTransE & - & - & - & \textbf{58} & - & 84.6 & - & - & - & - & - & -\tabularnewline
KG2E & 331 & - & 92.8 & 59 & - & 74.0 & - & - & - & - & - & -\tabularnewline
ManifoldE & - & - & 93.2 & - & - & 88.1 & - & - & - & - & - & -\tabularnewline
DistMult{*} & - & 0.83 & 93.6 & - & 0.35 & 57.7 & 5110 & 0.425 & 49.1 & 254 & 0.241 & 41.9\tabularnewline
ComplEx{*} & - & 0.941 & 94.7 & - & 0.69 & 84.0 & 5261 & 0.444 & \textbf{50.7} & 248 & 0.240 & 41.9\tabularnewline
ConvE & 504 & \textbf{0.942} & \textbf{95.5} & 64 & \textbf{0.745} & \textbf{87.3} & 7323 & 0.342 & 41.1 & 330 & \textbf{0.301} & 45.8\tabularnewline
\hline 
\end{tabular}
\end{table*}

Given a test triple $(h,r,t)$ with either $h$ or $t$ is missing,
our target is to complete this triple by finding the correct entity.
Similar to \cite{bordes2013translating}, we formularize this task
as a ranking problem. First, we replace the missing entity with every
entity in the knowledge graph and compute the energy of each candidate
triple in turn using Eq.~\ref{eq:TransH_energy}. Next, we filter
out all correct triples in the knowledge graph different from the
target one. Finally, we rank the energy values over the remaining
triples in ascending order and use those ranks to decide which entity
is the most suitable.

We used grid search for hyper-parameter tuning with the margin $\gamma$
is among $\{1,2,4\}$, the size of entity embedding $\drm_{e}$ and
relation embedding $\drm_{e}$ are among $\{20,50,100,150,200\}$,
the number of relation space bases $\srm$ is among $\{3,5,10,15\}$.
The optimizer is Adam \cite{kingma2014adam} with the learning rate
$\lambda$ of $0.001$. Following other papers, we pretrained our
model with TransE by setting the relation coefficients to $0$ for
$1000$ epochs then continued training for a maximum of $150$ epochs.
The optimal settings of our model on the validation set are $\gamma=4$,
$\drm_{e}=\drm_{r}=50$, $\srm=5$ for WN18 and WN18RR, $\gamma=2$,
$\drm_{e}=\drm_{r}=150$, $\srm=5$ for FB15k and $\gamma=4$, $\drm_{e}=\drm_{r}=100$,
$s=5$ for FB15k-237.

There are three evaluation metrics for this task: (i) \textit{Mean
Rank (MR)} (ii) \textit{Mean Reciprocal Rank (MRR)} and (iii) \textit{Hits@10}.
A better model would expect lower \textit{MR} and higher \textit{MRR/Hits@10}.
The overall results are shown in Table.~\ref{tab:link_pred}. TransF
outperforms all translation-based models by a large margin on WN18
and FB15k, achieving the best results on all evaluation metrics. Compared
to methods belonging to other disciplines, TransF also demonstrates
good performance with the best MR (3246) on WN18RR and the best Hits@10
(47.2\%) on FB15k-237. It suggests that factorizing the relation space
as a combination of multiple sub-spaces is critical for representing
different types of relations in knowledge graphs. This statement is
further supported when looking at Table.~\ref{tab:Detail_HEP_TEP}.
For 1-to-1 and N-to-N relations, TransF produces higher accuracy than
all baseline models in both HEP and TEP. Specifically, for HEP with
N-to-1 relations and TEP with 1-to-N relations, our model improves
the results by about 9\% and 5\% over the second best model TransSparse
(us), respectively.

\begin{table*}
\caption{\textit{Hits@10} results on FB15k arranged by different relation types.\label{tab:Detail_HEP_TEP}}

\centering{}%
\begin{tabular}{|c|c|c|c|c||c|c|c|c|}
\hline 
\multirow{2}{*}{Method} & \multicolumn{4}{c||}{Head Entity Prediction (HEP)} & \multicolumn{4}{c|}{Tail Entity Prediction (TEP)}\tabularnewline
\cline{2-9} 
 & 1-to-1 & 1-to-N & N-to-1 & N-to-N & 1-to-1 & 1-to-N & N-to-1 & N-to-N\tabularnewline
\hline 
\hline 
TransE & 43.7 & 65.7 & 18.2 & 47.2 & 43.7 & 19.7 & 66.7 & 50.0\tabularnewline
TransH & 66.8 & 87.6 & 28.7 & 64.5 & 65.5 & 39.8 & 83.3 & 67.2\tabularnewline
TransR & 78.8 & 89.2 & 24.1 & 69.2 & 79.2 & 37.4 & 90.4 & 72.1\tabularnewline
CTransR & 81.5 & 89.0 & 34.7 & 71.2 & 80.8 & 38.6 & 90.1 & 73.8\tabularnewline
TransD & 86.1 & 95.5 & 39.8 & 78.5 & 85.4 & 50.6 & 94.4 & 81.2\tabularnewline
TransSparse (s) & 86.8 & 95.5 & 44.3 & 80.9 & 86.6 & 56.6 & 94.4 & 83.3\tabularnewline
TransSparse (us) & 87.1 & \textbf{95.8} & 44.4 & 81.2 & 87.5 & 57.0 & \textbf{94.5} & 83.7\tabularnewline
\hline 
TransF & \textbf{88.1} & 94.9 & \textbf{53.2} & \textbf{82.8} & \textbf{88.8} & \textbf{62.1} & 93.4 & \textbf{85.8}\tabularnewline
\hline 
\end{tabular}
\end{table*}

\subsection{Relation Representation with TransF.}

\begin{figure*}
\begin{centering}
\includegraphics[scale=0.29]{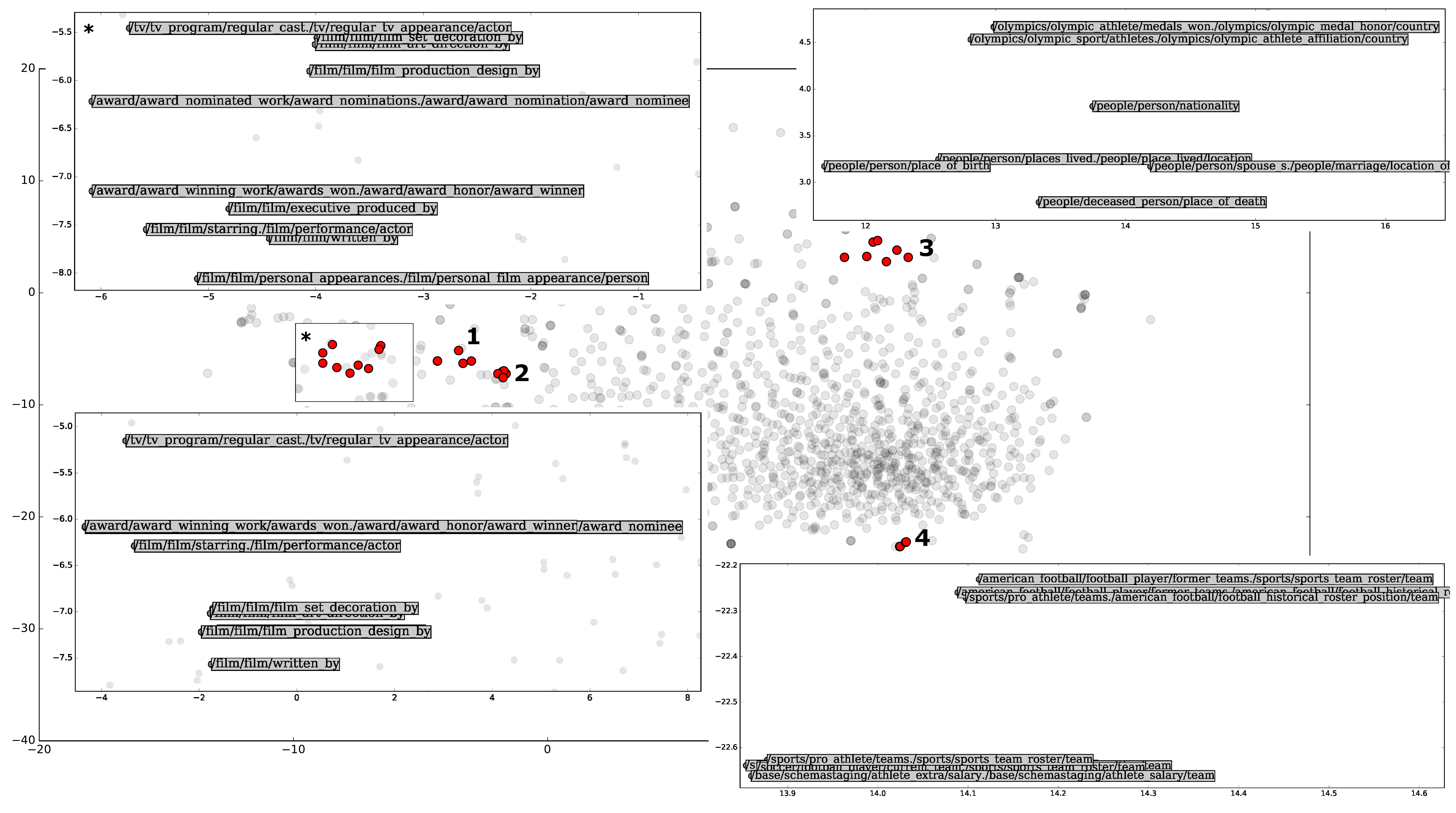}
\par\end{centering}
\caption{tSNE visualization of all relations in FB15k. For each relation $r$,
the representation vector is constructed by concatenating the translation
vector $\protect\rbm$ and two coefficient vectors $\protect\alphab_{r}$
and $\protect\betab_{r}$ into a single vector. Some related relations
are highlighted in red and are grouped together. To better understand
the meaning of these relations, they are zoomed with names. The small
rectangle snapshot marked with {*} is the tSNE embedding of relations
in group $1$ and $2$ by using the translation vector $\protect\rbm$
only.\label{fig:tSNE-visualization}}
\end{figure*}

In Fig.~\ref{fig:tSNE-visualization}, we show the t-SNE visualization
\cite{maaten2008visualizing} of all relations in FB15. There are
three main things to note here: (i) Our model successfully captures
the correlations among relations as relations with similar semantic
meaning usually stay close in the embedded space. (ii) Correlated
relations do not necessarily belong to the same category but can span
across different categories. For example, in group 4 two relations
``/base/schemastaging/.../team'' and ``/sports/pro\_athlete/.../team''
are from two categories ``base'' and ``sports'', respectively.
(iii) Although most information of a relation is stored in the translation
vector due to its large size, the coefficient vectors also provide
certain amount of information to make the relation representation
more accurate. As in Fig.~\ref{fig:tSNE-visualization}, when relation
representation does not contain coefficient vectors, the relations
in group 1 and 2 seem to merge together even though they are not very
similar. On the other hand, when coefficient vectors are used, these
two groups are more separately.

\subsection{Complexity Analysis of TransF.}

\begin{figure}
\begin{centering}
\includegraphics[scale=0.25]{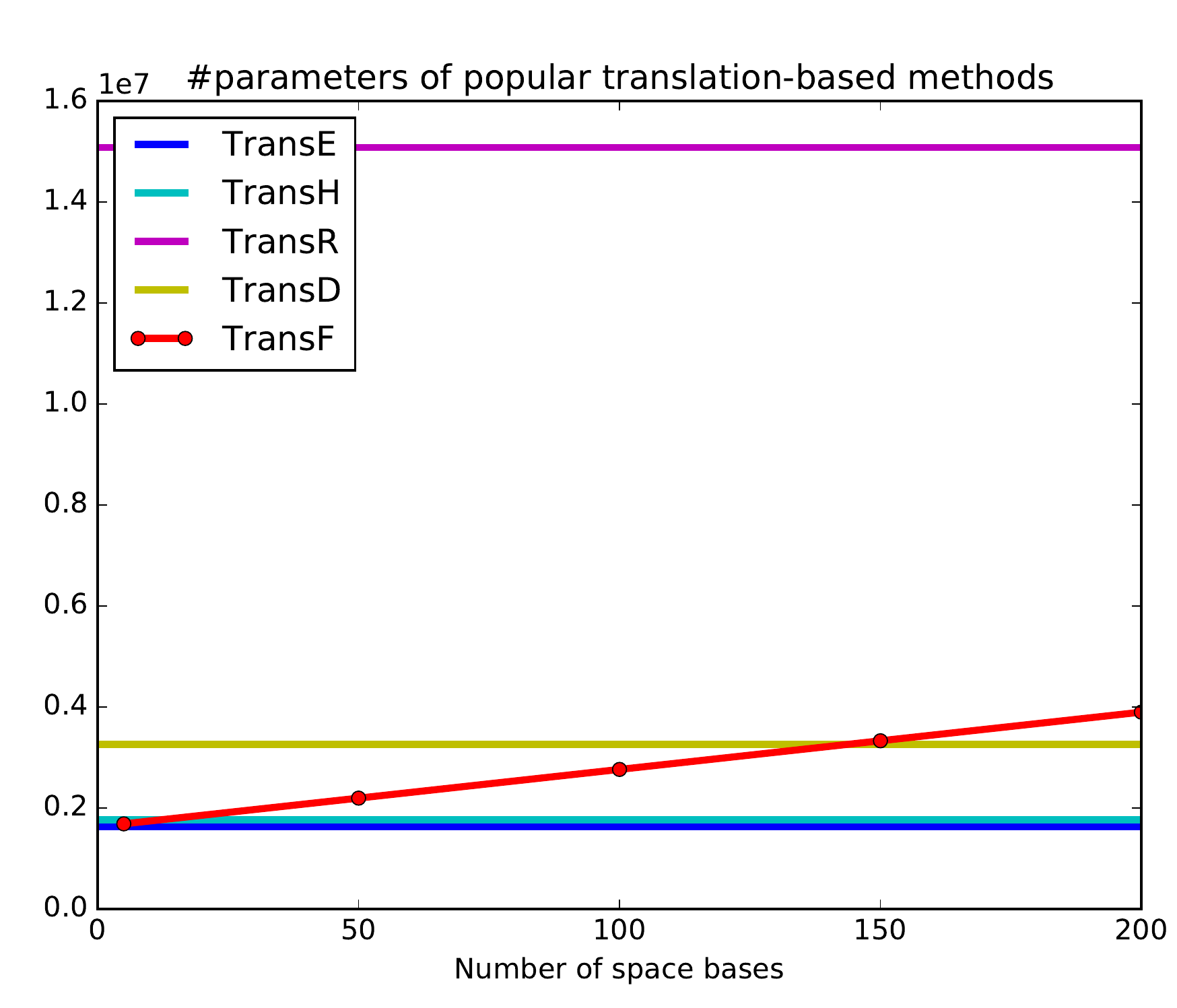}\includegraphics[scale=0.25]{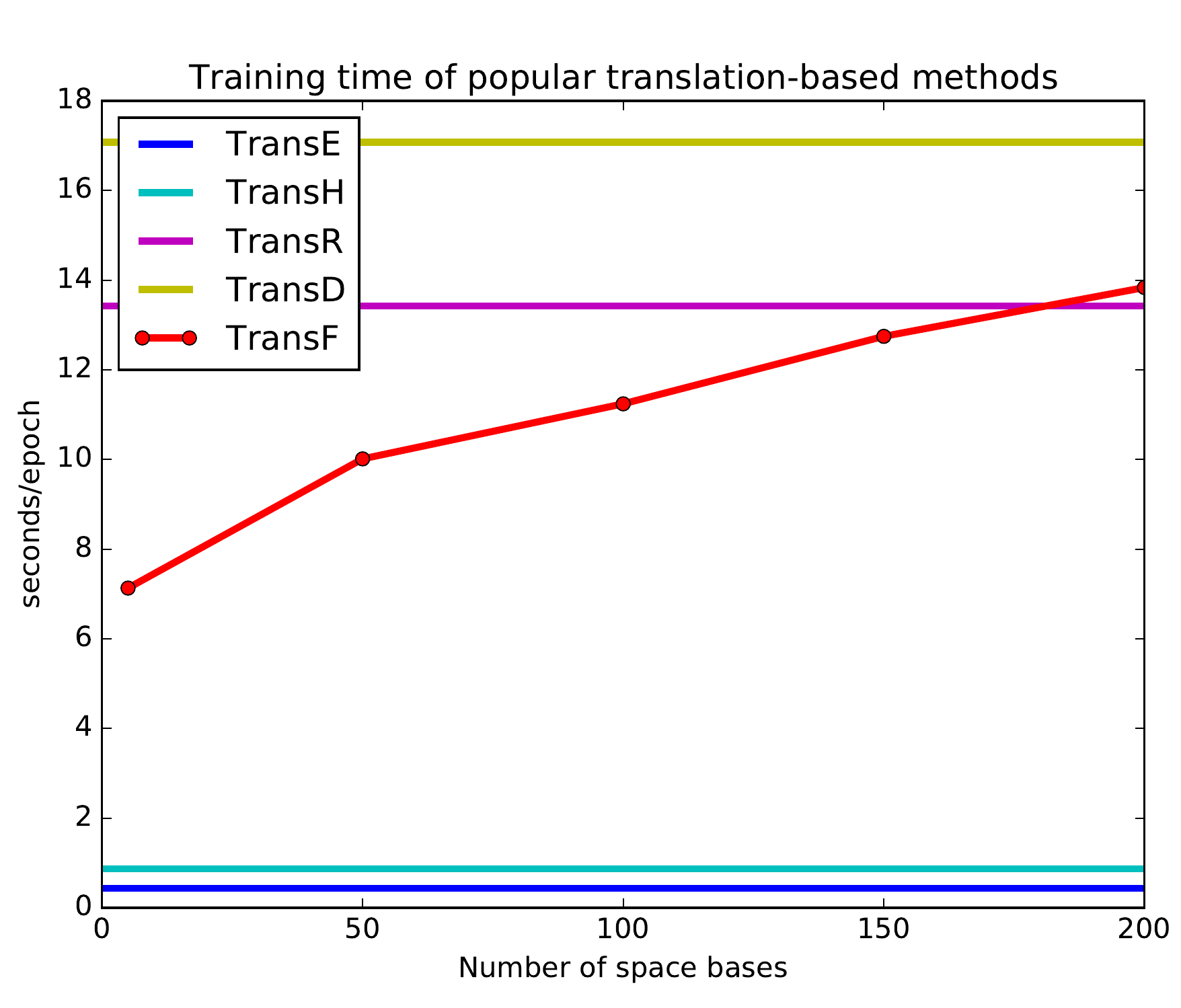}
\par\end{centering}
\caption{The number of parameters (left) and training time (right) of popular
translation-based methods on FB15k. In all methods, the entity and
relation embedded dimensions are both set to $100$. The training
time is recorded for $50$ epochs and then taking average. Models
are programmed by authors using Theano and run on a single GPU GTX
980Ti 12GB.\label{fig:params_training}}
\end{figure}

In Fig.~\ref{fig:params_training} we plot the number of parameters
and training time of TransF in comparison with some other translation-based
methods on FB15k. When the number of space bases $\srm$ is small,
TransF has nearly the same number of parameters as TransE and TransH.
Increasing $\srm$ only grows the number of parameters by a small
constant rate. Not only consuming fewer parameters, TransF also runs
faster than TransR and TransD. Specifically, with s=5, TransF can
be trained in roughly half amount of time compared to TransR. This
difference reduces to about 20\% when $\srm=100$. In this experiment,
we also observe an unexpected pattern: TransD is slower than TransR
though its projection matrix is factorized as product of two vectors.
This is because TransD has to recompute its projection matrix for
every triple while TransR can index its projection matrix based on
relations. Our method does not suffer this problem, hence, is much
more efficient than TransD. However, the training time of TransF is
still not comparable to TransE and TransH. It explains why pretraining
with TransE is necessary.

\section{Related Work}

\subsubsection*{Translation based methods}

Beside Trans(E, H, R, D, Sparse) that we have already discussed, there
are other models falling into this category. \textbf{lppTransD} \cite{yoon2016translation}
is an extension of TransD that accounts for different roles of head
and tail entities. They showed that logical properties like transitivity
and symmetry cannot be represented by using the same projection matrix
for both head and tail entities. This idea is also applied in our
model as we use two separate sets of basis matrices ($\{\Ubm^{(i)}|i=\overline{1,\srm}\}$
and $\{\Vbm^{(i)}|i=\overline{1,\srm}\}$) to compute the projection
matrices for head and tail entities. \textbf{STransE} \cite{nguyen2016stranse}
combines Structured Embedding (SE) \cite{bordes2011learning} and
TransE into a single model. Its energy function is $E(h,r,t)=\|\Wbm_{r,1}\hbm+\rbm-\Wbm_{r,2}\tbm\|_{\ell_{1/2}}$.
In fact, this model is similar to lppTransR \cite{yoon2016translation}.
\textbf{KB2E} \cite{he2015learning} takes an interesting approach
to handle non-injective relations by incorporating knowledge graph
uncertainty into embedding. Specifically, it models entities and relations
as Gaussian distributions instead of single points in the embedding
space: $x\sim\Ncal(\boldsymbol{\mu}_{x},\boldsymbol{\Sigma}_{x})$
for $x=h$, $t$, $r$. The KL divergence between two distributions
$h-t$ and $r$ is selected to be an (asymmetric) energy function
very naturally: $E=D_{KL}(h-t,r)=D_{KL}(\Ncal(\boldsymbol{\mu}_{h}-\boldsymbol{\mu}_{t},\boldsymbol{\Sigma}_{h}+\boldsymbol{\Sigma}_{t}),\Ncal(\boldsymbol{\mu}_{r},\boldsymbol{\Sigma}_{r}))$.

\subsubsection*{Tensor based methods}

Tensor based methods represent a knowledge graph as a 3D tensor $\Xcal$
of shape $\Nrm_{e}\times\Nrm_{e}\times\Nrm_{r}$ where $\Nrm_{e}$
and $\Nrm_{r}$ are the number of entities and relations in the knowledge
graph, respectively. Each element $\Xcal_{i,j,k}$ of the tensor can
be seen as probability that the triple $(e_{i},r_{k},e_{j})$ is correct.

\textbf{RESCAL} \cite{nickel2011three} applies tensor factorization
to estimate $\Xcal$. Specifically, each slice matrix $\Xcal_{:,:,k}$
($k=\overline{1,\Nrm_{r}}$) along the relation axis is computed as
$\Xcal_{:,:,k}=\Ebm\Rbm_{k}\Ebm^{\intercal}$where $\Ebm\in\Real^{\Nrm_{e}\times\srm}$
is a latent factor matrix of the entities, $\Rbm_{k}\in\Real^{\srm\times\srm}$
is an matrix that models the interactions of the components with respect
to the $k$-th relation. Compared to TransE with the same number of
hidden units, RESCAL requires far more parameters (as much as TransR).
In addition, the three-way dot product make this model more difficult
to be trained. This is the reason why RESCAL is not comparable to
TransE in many situations \cite{bordes2013translating}.

\textbf{DistMult} \cite{yang2014embedding} is a simplified version
of RESCAL with the energy function $E(h,r,t)=\hbm^{\intercal}\Wbm_{r}^{\text{diag}}\tbm$.
Here, the interaction between head and tail entities is captured via
a diagonal matrix $\Wbm_{r}^{\text{diag}}\in\Real^{\drm_{e}\times\drm_{e}}$
instead of a 3D tensor like in NTN. In fact, we can rewrite the energy
function of DistMult as $E(h,r,t)=\text{sum}(\hbm\odot\rbm\odot\tbm)$
where $\rbm\in\Real^{\drm_{e}}$ is the main diagonal of $\Wbm_{r}^{\text{diag}}$.
In this form, DistMult looks very similar to TransE but with additive
operators replaced by multiplicative ones.

\textbf{Holographic Embedding (HolE)} \cite{nickel2016holographic},
a novel method leveraging the holographic models of associative memory
to learn the compositional representations of knowledge graphs. The
probability of a triple $(e_{i},r_{k},e_{j})$ to be correct is computed
as: $\Xcal_{i,j,k}=\sigma(\phi_{i,j,k})=\sigma(\rbm_{k}^{\intercal}(\ebm_{i}\star\ebm_{j}))$
where $\phi_{i,j,k}$ is a characteristic function over the triple,
$\rbm_{k}\in\Real^{\drm_{r}}$, $\ebm_{i},\ebm_{j}\in\Real^{\drm_{e}}$
are the relation and entity embeddings; $\star$ is a circular correlation
operator. From the holography angle \cite{poggio1973holographic},
we can see that the association of $\rbm_{k}$ and $\ebm_{i}$ is,
first, implicitly stored in $\ebm_{j}$ via training. Then, taking
a circular correlation with $\ebm_{i}$ will return $\rbm'_{k}$ -
a noisy version of $\rbm_{k}$. And finally, the dot product with
$\rbm_{k}$ will examine how similar this two vectors are. HolE has
many advantages such that computation efficiency (only calculating
on vectors), scalability (the number of parameters is small) and capability
of representing anti-symmetric relations ($\star$ is non-commutative).

Another model closely related to HolE is \textbf{Complex Embedding
(ComplEx)} \cite{trouillon2016complex}. It originates from the observation
that embedding relations and entities into complex spaces would be
better than into real spaces due to the non-symmetry of the Hermitian
product. The characteristic function $\phi_{i,j,k}$ of ComplEx is
defined as: $\phi_{i,j,k}=\text{Re}(<\rbm_{k},\ebm_{i},\overline{\ebm}_{j}>)$
where $\rbm_{k},\ebm_{i},\ebm_{j}$ are complex vectors; $<a,b,c>$
is a trilinear product between $a,b,c$; $\overline{a}$ is the conjugate
of $a$. ComplEx and HolE have been proven to be mathematically equivalent
\cite{trouillon2017complex}. Thus, both models provide the same representation
power.

\subsubsection*{Other related methods}

One early work that applied the embedding concept for knowledge graph
completion is \textbf{Structured Embedding (SE)} \cite{bordes2011learning}.
 The basic idea of this model is that two entities of a correct triple
should be close to each other in some relation spaces. Hence, its
energy function is defined as $E(h,r,t)=\|\Wbm_{r,1}\hbm-\Wbm_{r,2}\tbm\|_{\ell_{1/2}}$. 

\textbf{Semantic Matching Energy (SME)} \cite{bordes2014semantic}
introduces the relation embedding $\rbm$ and treats it equally to
the entity embeddings $\hbm$ and $\tbm$. This method is suitable
for situations when relations and entities are interchangable, for
example, in NLP, a verb typically corresponds to a relation but sometimes
can also be an entity. SME defines the energy function as $E(h,r,t)=-f(\hbm,\rbm)^{\intercal}g(\tbm,\rbm)$
where $f$ and $g$ are neural networks. In case $f$ and $g$ are
linear, this energy function only captures two-way interactions of
pairs $(h,r)$, $(t,r)$ and $(t,h)$ rather than the three-way interaction
as in RESCAL. . 

Both SE and RESCAL can be seen as special cases of \textbf{Neural
Tensor Network (NTN)} whose energy function is $E(h,r,,t)=\ubm_{r}^{\intercal}\sigma(\hbm^{\intercal}\Wbcal_{r}\tbm+\Ubm_{r}\hbm+\Vbm_{r}\tbm+\bbm)$
where $\sigma$ is a nonlinear activation function (e.i. $\tanh$),
$\Wbcal_{r}\in\Real^{\drm_{e}\times\drm_{e}\times k}$ , $\Ubm_{r},\Vbm_{r}\in\Real^{\drm_{e}\times k}$
and $\ubm_{r}\in\Real^{k}$ all depends on $r$. Despite being expressive,
NTN does not scale well to knowledge graphs with large number of relations
(e.g. FB15k) due to its high computational cost.

\textbf{}

\section{Conclusion}

We have proposed TransF, a new knowledge graph embedding method explicitly
models the relationship between relations. Our model decomposes the
relation\textendash specific projection spaces into a small number
of spanning bases, which are shared by all relations. We showed that
this strategy not only leads to better performance but is also more
efficient than state-of-the-art translation-based methods like TransR
or TransD through extensive experiments on link prediction and complexity
analysis. In addition, the visualization of learnt relations also
indicates that TransF models the relation correlations well. In the
future, we plan to explore better representation of relations. One
potential way is using additional information from larger structures
such as paths or subgraphs instead of triples only. Another direction
is to define class\textendash specific model of relations.

\bibliographystyle{plain}

\end{document}